\documentclass{hld2024} 

\usepackage{xcolor}
\usepackage{amsmath}
\usepackage{amssymb}
\usepackage{mathtools}
\usepackage{booktabs}
\usepackage{enumitem}

\renewcommand{\vec}{\mathbf}
\renewcommand{\matrix}{\mathbf}

\title[Physics-Guided Policy Optimization with Self-Distillation]{Physics-Guided Policy Optimization with Self-Distillation}

\hldauthor{%
\Name{Ke Wang$^{*}$} \Email{kewangv@amazon.com}\\
\addr {Amazon, Arlington, USA} 
\AND
\Name{Yuning Wu$^{*}$} \Email{yuningwu@amazon.com}\\
\addr {Amazon, New York, USA}
\AND
\Name{Haoran Liu} \Email{liuhr@amazon.com}\\
\addr {Amazon, Arlington, USA}
\AND
\Name{Chaoqun Jia} \Email{enzojia@amazon.com}\\
\addr {Amazon, Arlington, USA}
\AND
\Name{Devin Chen} \Email{devichen@amazon.com}\\
\addr {Amazon, San Jose, USA}
\AND
\Name{Kai Wei} \Email{kaiwe@amazon.com}\\
\addr {Amazon, Seattle, USA}
}

\begin{document}

\maketitle
\renewcommand{\thefootnote}{\fnsymbol{footnote}}%
  \footnotetext[1]{Equal contribution.}%
  \renewcommand{\thefootnote}{\arabic{footnote}}

\begin{abstract}%

\noindent Self-distilled policy optimization (SDPO) has become a popular paradigm for LLM post-training, where a model learns from its own predictions conditioned on privileged information. SDPO, however, is sensitive to how much each update step should be trusted: corrections from a self-teacher can be highly informative on some batches and misleading on others, and applying them uniformly with a fixed step size can destabilize training. Drawing inspiration from viscous-fluid dynamics and formalizing the analogy at the SDE level, we propose Physics-Guided Policy Optimization (PGPO), which introduces an information-modulated step-size multiplier derived from a mutual-information estimate between the student's predictions and the feedback-conditioned teacher. We show that this modulation preserves the order-1 weak-approximation guarantees of vanilla SGD, and incurs negligible overhead per iteration. We evaluate PGPO on the Science-QA dataset, where it outperforms SDPO on 3 of the 4 domains with gains of up to +4.5 points, while remaining stable in a setting where SDPO collapses late in training.

\end{abstract}


\begin{figure}[h]
    \centering
    \includegraphics[width=0.85\linewidth]{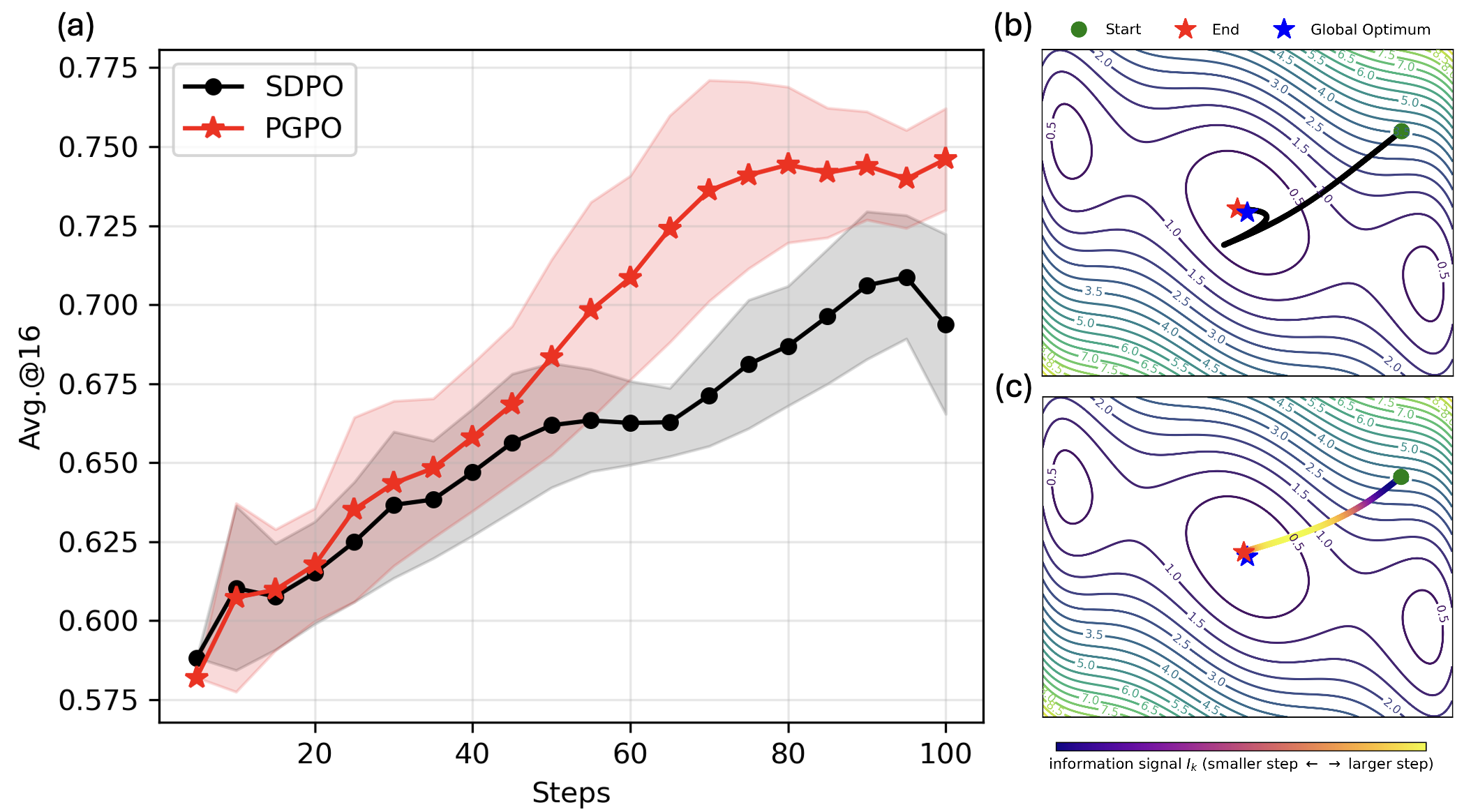}
    \caption{(a) Avg.@16 of PGPO vs. SDPO on Science Q\&A (Physics). Motivating intuition: (b) SDPO applies a uniformly-scaled step; (c) PGPO scales each step by an information signal $I_k$ to an oracle teacher (illustration only) with brighter colors marking larger steps.}
    \label{fig:viscosity-comparison}
\end{figure}

\section{Introduction}


\noindent On-policy distillation (OPD)~\citep{lu2025onpolicydistillation,song2026surveyonpolicydistillationlarge,li2026rethinkingonpolicydistillationlarge,coreteam2026mimov2flashtechnicalreport,deepseekai2026deepseekv4} has emerged as a popular training paradigm for large language model post-training, utilizing a teacher model to generate dense, token-level guidance for supervising student model training. A recent variant, self-distilled policy optimization (SDPO)~\citep{hübotter2026reinforcementlearningselfdistillation}, has attracted community attention by treating a single model as both teacher and student: the teacher receives privileged information (e.g., correct solutions) to generate signals, while the student learns from these signals without such access. This method eliminates the need for a separate, larger teacher model and facilitates iterative self-improvement. Existing studies~\citep{shenfeld2026selfdistillationenablescontinuallearning,song2026expandingcapabilitiesreinforcementlearning,zhao2026selfdistilledreasoneronpolicyselfdistillation,ye2026onpolicycontextdistillationlanguage,kim2026doesselfdistillationsometimesdegrade} have shown that combining SDPO with reinforcement learning with verified rewards (RLVR)~\citep{lambert2024tulu} leads to efficient performance gains.

\noindent However, SDPO faces critical challenges during training. While dense token-level supervision enables rapid early gains, it becomes unstable as the student model evolves. As generated prefixes drift from the teacher's conditioning distribution, corrections become unreliable or directionally misleading. Moreover, applying corrections uniformly across all samples, including correct ones, introduces ambiguity and corrupts learned behaviors. Extreme token-level rewards further destabilize training by dominating gradient updates, ultimately leading to collapse. Behind these symptoms lies a fundamental credit assignment problem: \textit{treating all tokens and samples uniformly fails to account for their varying informativeness and reliability.}

\noindent To alleviate these issues, several methodologies have been proposed: routing corrections based on sample correctness~\citep{li2026unifyinggrouprelativeselfdistillationpolicy}, filtering signals by model confidence~\citep{yang2026selfdistilledrlvr}, and anchoring updates to environment rewards~\citep{ko2026scalingreasoningefficientlyrelaxed}. In contrast to these approaches, we propose \textbf{Physics-Guided Policy Optimization} (PGPO), a method grounded in physical intuition from fluid mechanics~\citep{cengel2013ebook}. PGPO interprets the optimization process as a particle moving through a viscous medium, where viscosity adaptively modulates step sizes based on local mutual information. This mechanism mirrors non-Newtonian fluids, where viscosity adapts to flow conditions, providing a principled alternative to uniform correction strategies that preserves informative signals while suppressing noise. By scaling each update step according to the informativeness of the current self-teacher signal, PGPO addresses the credit assignment problem and prevents the instability that plagues training.

\noindent Our contributions are as follows:
\begin{itemize}
    \item We propose Physics-Guided Policy Optimization (PGPO), a method inspired by fluid mechanics that addresses credit assignment in self-distilled policy optimization (SDPO).
    \item We show that PGPO admits an order-1 weak SDE approximation and demonstrate its effectiveness in preliminary experiments.
\end{itemize}

\section{Preliminaries}

\subsection{Problem Formulation}

Consider a dataset $\mathcal{D} = \{(\vec{x}_i, \vec{y}_i)\}_{i=1}^{N}$ consisting of $N$ prompt-response pairs, where $\vec{x}_i$ denotes the input prompt and $\vec{y}_i$ the target response from offline datasets or online rollouts. We aim to optimize a large language model (LLM) parameterized by $\vec{\theta} \in \mathbb{R}^d$, denoted as $\pi_{\vec{\theta}}$, by solving the empirical risk minimization problem:

\begin{equation}
\min_{\vec{\theta} \in \mathbb{R}^d} \mathcal{L}(\vec{\theta}) := \frac{1}{N} \sum_{i=1}^{N} \ell_i(\vec{\theta}),
\end{equation}

\noindent where $\ell_i(\vec{\theta}) := \ell(\vec{\theta}; \vec{x}_i, \vec{y}_i)$ represents the training objective on the $i$-th example. We use this notation throughout the paper for simplicity. 

\noindent In self-distilled policy optimization (SDPO)~\citep{hübotter2026reinforcementlearningselfdistillation}, the student $\pi_{\vec{\theta}}$ generates responses, receives feedback $f_i$, and distills from a self-teacher that retrospectively corrects mistakes:

\begin{equation}
\ell_i(\vec{\theta}) = \sum_{t} \mathbb{D}_{\text{KL}}\left( \pi_{\vec{\theta}}(\cdot | \vec{x}_i, \vec{y}_{<t}) \,\|\, \text{stopgrad}(\pi_{\vec{\theta}}(\cdot | \vec{x}_i, f_i, \vec{y}_{<t})) \right).
\end{equation}

\subsection{Stochastic Gradient Descent}

\noindent To minimize $\mathcal{L}(\vec{\theta})$, we can employ stochastic gradient descent (SGD) and its adaptive variants (e.g., Adam~\citep{kingma2017adammethodstochasticoptimization}). At each iteration $k$, SGD samples a mini-batch $\mathcal{B}_k \subset \{1, ..., N\}$ and updates with learning rate $\eta > 0$:

\begin{equation}
\vec{\theta}_{k+1} = \vec{\theta}_k - \eta \cdot \nabla \widehat{\mathcal{L}}(\vec{\theta}_k), \quad \text{where} \quad \nabla \widehat{\mathcal{L}}(\vec{\theta}_k) = \frac{1}{|\mathcal{B}_k|} \sum_{i \in \mathcal{B}_k} \nabla_{\vec{\theta}} \ell_i(\vec{\theta}_k).
\label{eq:sgd_update}
\end{equation}

\subsection{Stochastic Differential Equation}

The dynamics of SGD can be interpreted as stochastic differential equations (SDEs)~\citep{li2017stochasticmodifiedequationsadaptive}. Rewrite the SGD update rule as:
\begin{equation}
\vec{\theta}_{k+1} - \vec{\theta}_k = -\eta \cdot \nabla \mathcal{L}(\vec{\theta}_k) + \sqrt{\eta} \cdot \vec{v}_k,
\end{equation}
where $\vec{v}_k = \sqrt{\eta} \cdot (\nabla \mathcal{L}(\vec{\theta}_k) - \nabla \widehat{\mathcal{L}}(\vec{\theta}_k))$ is a $d$-dimensional random vector representing the gradient noise. By the Central Limit Theorem~\citep{wasserman2004all}, for sufficiently large mini-batch sizes, $\vec{v}_k$ is approximately Gaussian with mean zero and covariance matrix $\eta \cdot \matrix{\Sigma}(\vec{\theta}_k)$, where
\begin{equation}
\label{eq:covar}
\matrix{\Sigma}(\vec{\theta}) = \frac{1}{N} \sum_{i=1}^{N} \left(\nabla \mathcal{L}(\vec{\theta}) - \nabla_{\vec{\theta}} \ell_i(\vec{\theta})\right) \, \left(\nabla \mathcal{L}(\vec{\theta}) - \nabla_{\vec{\theta}} \ell_i(\vec{\theta})\right)^\top
\end{equation}
captures the variance of stochastic gradients across dataset.

\noindent To derive the continuous-time limit, we identify the discrete update with the Euler-Maruyama discretization of an SDE~\citep{kloden1992numerical}. Setting the time step $\Delta t = \eta$ and recognizing that $\vec{v}_k \approx \sqrt{\eta} \cdot \matrix{\Sigma}^{1/2}(\vec{\theta}_k) \, \vec{z}_k$, where $\vec{z}_k \sim \mathcal{N}(\vec{0}, \matrix{I}_d)$ and $\matrix{\Sigma}^{1/2}$ denotes the matrix square root. 
\begin{equation}
\vec{\theta}_{k+1} = \vec{\theta}_k - \eta \cdot \nabla \mathcal{L}(\vec{\theta}_k) + \eta \cdot \matrix{\Sigma}^{1/2}(\vec{\theta}_k) \, \vec{z}_k.
\end{equation}

\noindent Taking the continuous-time limit yields the SGD-SDE:
\begin{equation}
d\vec{\theta}_t = \underbrace{-\nabla \mathcal{L}(\vec{\theta}_t)}_{\text{drift}} \, dt + \sqrt{\eta} \cdot \underbrace{\matrix{\Sigma}^{1/2}(\vec{\theta}_t)}_{\text{diffusion}} \, d\vec{w}_t.
\label{eq:SDE}
\end{equation}
where $\vec{w}_t$ is a standard $d$-dimensional Brownian motion. The learning rate $\eta$ sets the time scale: one unit of SDE time corresponds to $1/\eta$ SGD iterations. This SGD-SDE provides a continuous-time approximation of SGD in the \textit{weak sense}: their statistical properties (e.g., marginal distributions, expectations) converge as $\eta \to 0$ (see Appendix~\ref{sec:weak_convergenc_proof}).

\begin{algorithm2e}
\caption{Physics-Guided Policy Optimization (PGPO)}
\label{alg:pgpo}
\SetAlgoLined
\KwData{Dataset $\mathcal{D} = \{(\vec{x}_i, \vec{y}_i)\}_{i=1}^{N}$, initial $\vec{\theta}_0$, learning rate $\eta$, sensitivity $\alpha$, max multiplier $\rho_{\max}$, iterations $K$}
\KwResult{Optimized parameters $\vec{\theta}_K$}
Initialize optimizer state\;
\For{$k = 0, 1, \ldots, K-1$}{
    Sample mini-batch $\mathcal{B}_k \subset \{1, \ldots, N\}$\;
    
    Generate responses and collect feedback for SDPO\;

    Compute gradient: $\vec{g}_k = \frac{1}{|\mathcal{B}_k|} \sum_{i \in \mathcal{B}_k} \nabla_{\vec{\theta}} \ell_i(\vec{\theta}_k)$\;
    
    Compute information signal $I_k$\;
    
    Compute physics-guided multiplier: $\rho_k = \min( \exp( \alpha \cdot I_k ), \rho_{\max} )$\;
    
    Update: $\vec{\theta}_{k+1} = \vec{\theta}_k - \eta \cdot \rho_k \cdot \vec{g}_k$\;
}
\Return{$\vec{\theta}_K$}\;
\end{algorithm2e}

\section{Physics-Guided Policy Optimization}

\noindent The SGD-SDE can be interpreted through fluid mechanics~\citep{cengel2013ebook}, where the parameter trajectory $\vec{\theta}_t$ represents a particle moving through a viscous fluid. The drift term $-\nabla \mathcal{L}(\vec{\theta}_t)$ acts as a pressure gradient driving the particle toward loss minima, while the diffusion term $\matrix{\Sigma}^{1/2}(\vec{\theta}_t) d\vec{w}_t$ captures stochastic fluctuations from mini-batch sampling, analogous to Brownian motion. This perspective suggests a natural mechanism for adaptive step-size control: modulating viscosity based on how informative the self-teacher's correction is. In physical systems, high viscosity slows motion while low viscosity permits faster movement; we let an uninformative correction keep baseline viscosity, and an informative correction reduce it.


\noindent We quantify the informativeness of the correction using the mutual information between predictions and feedback on mini-batch $\mathcal{B}_k$:

\begin{equation}
\label{eq:uq}
I_k := \frac{1}{|\mathcal{B}_k|} \sum_{i \in \mathcal{B}_k} \left[ H(\vec{y}_i | \vec{x}_i, \vec{y}_{<t}) - H(\vec{y}_i | \vec{x}_i, f_i, \vec{y}_{<t}) \right],
\end{equation}

\noindent where high $I_k$ indicates that the feedback is highly informative; the student's prediction without feedback diverges from the feedback-conditioned teacher, so the correction carries a strong, reliable signal. Conversely, low $I_k$ indicates that the student already agrees with the teacher, making the correction small. This motivates an adaptive viscosity that decreases (taking larger steps) when the correction is informative, and stays at baseline when it is not.

\noindent With the physical interpretation and a quantified informativeness signal in hand, we introduce Physics-guided Policy Optimization (PGPO), where a multiplier $\rho(I_k)$ modulates the gradient step:
\begin{equation}
\vec{\theta}_{k+1} = \vec{\theta}_k - \eta \cdot \rho(I_k) \cdot \nabla \widehat{\mathcal{L}}(\vec{\theta}_k), \quad \text{where} \quad \rho(I_k) = \min\left( \exp\left( \alpha \cdot I_k \right), \rho_{\max} \right),
\end{equation}
with $\alpha > 0$ controlling sensitivity and $\rho_{\max} \geq 1$ capping acceleration. This yields:
\begin{itemize}
    \item High $I_k$ (informative correction): $\rho_k > 1$ → reduced viscosity → larger steps
    \item Low $I_k$ (uninformative correction): $\rho_k \approx 1$ → baseline viscosity → standard steps
\end{itemize}

\noindent The cap $\rho_{\max}$ mirrors a physical safety constraint, limiting acceleration when the information signal spikes, while $\rho_k \geq 1$ keeps each update at least as large as the baseline SGD step.


\noindent Algorithm~\ref{alg:pgpo} presents the complete PGPO framework. The key innovation is the physics-guided multiplier $\rho_k$, which modulates step size based on the per-batch information signal $I_k$, with only $O(1)$ overhead per iteration, since $I_k$ reuses the forward pass used for gradient computation. The framework extends to adaptive optimizers (e.g., Adam~\citep{kingma2017adammethodstochasticoptimization}) by modulating the final update without altering moment accumulation: for Adam with update direction $\vec{h}_k$, the viscosity-modulated update becomes $\vec{\theta}_{k+1} = \vec{\theta}_k - \eta \cdot \rho_k \cdot \vec{h}_k$.

\subsection{Experimental Setup}

\noindent \textbf{Models.} We use Qwen3-8B~\citep{yang2025qwen3technicalreport}, the general reasoning model, as the initial checkpoint.

\noindent \textbf{Training.} We train on Science Q\&A~\citep{feng2025sciknowevalevaluatingmultilevelscientific} across four domains (Chemistry, Physics, Biology, Materials Science) using undergraduate-level reasoning problems, where each domain is a separate training task. We use sensitivity parameters $\alpha \in \{0.5, 1.0, 1.5\}$, with all other hyperparameters following~\citet{hübotter2026reinforcementlearningselfdistillation}, details in Appendix \ref{sec:experiment_details}.

\noindent \textbf{Evaluation.} We assess in-domain generalization on held-out test splits and report avg@16. All hyperparameters follow~\citet{hübotter2026reinforcementlearningselfdistillation}, details in Appendix \ref{sec:experiment_details}.

\subsection{Main Results}

\begin{table}[ht]
\centering
\caption{Performance comparison between SDPO and PGPO across scientific domains.}
\label{tab:results}
\begin{tabular}{lcccc}
\toprule
Method & Chemistry & Physics & Biology & Materials \\
\midrule
SDPO & 72.57 & 77.05 & 61.62 & 74.14 \\
PGPO (best $\alpha$) & 76.02 & 77.56 & 61.25 & 78.65 \\
\midrule
Improvement & +3.45 & +0.51 & -0.37 & +4.51 \\
\bottomrule
\end{tabular}
\end{table}

\noindent PGPO outperforms SDPO in 3 out of 4 scientific domains, achieving notable improvements in Chemistry (+3.45) and Materials (+4.51), with a modest gain in Physics (+0.51). However, PGPO shows a slight decline in Biology (-0.37). The results demonstrate that physics-guided adaptive step-size control provides consistent benefits across most domains, with the largest gains in Chemistry and Materials Science. The full ablations are presented in Appendix \ref{sec:ablations}.

\section{Conclusion}

We proposed Physics-Guided Policy Optimization (PGPO), a method that addresses the credit assignment problem in self-distilled policy optimization (SDPO) through adaptive step-size modulation motivated by an analogy with fluid mechanics. We showed that PGPO is an order-1 weak approximation of a modulated-drift SDE, and demonstrated through preliminary experiments that PGPO achieves consistent improvements over SDPO across multiple scientific reasoning domains. The current instantiation modulates step size at the batch level and is sensitive to the choice of sensitivity parameter $\alpha$, which we observed in the Biology setting where no single $\alpha$ matched SDPO; finer-grained (e.g., per-token) modulation and adaptive $\alpha$ schedules are natural directions for future work. These results suggest that PGPO offers an alternative approach to stabilizing self-distillation in large language model post-training.

\bibliography{sample}

\newpage
\appendix

\section{Weak Convergence Guarantees}
\label{sec:weak_convergenc_proof}

\noindent We now show that PGPO admits an order-1 weak approximation by an SDE with a modulated drift. We emphasize that this is a consistency-style result rather than a convergence-rate guarantee.

\begin{definition}[Weak Approximation]~\citep{mannella1997numericalintegrationstochasticdifferential}
\label{def:weak_approx}
Let $0 < \eta < 1$, $T > 0$, and set $N = \lfloor T/\eta \rfloor$. Let $\mathcal{G}$ denote the set of functions with polynomial growth, i.e., $g \in \mathcal{G}$ if there exist constants $K, \kappa > 0$ such that $|g(\vec{x})| \leq K(1 + \|\vec{x}\|^\kappa)$. We say that the SDE
\begin{equation}
d\vec{\theta}_t = \vec{b}(\vec{\theta}_t) \, dt + \matrix{\sigma}(\vec{\theta}_t) \, d\vec{w}_t, \quad \vec{\theta}_0 = \vec{\theta}_{\text{init}}
\end{equation}
is an \textbf{order $\alpha$ weak approximation} to SGD if for every $g \in \mathcal{G}$, there exists $C > 0$ independent of $\eta$ such that for all $k = 0, 1, \ldots, N$,
\begin{equation}
\left| \mathbb{E}[g(\vec{\theta}_{k\eta})] - \mathbb{E}[g(\vec{\theta}_k^{\text{SGD}})] \right| \leq C \eta^\alpha.
\end{equation}
\end{definition}

\begin{theorem}[Weak Convergence of PGPO]
\label{thm:weak_convergence}
Let $T > 0$ and define the gradient covariance $\matrix{\Sigma}(\vec{\theta})$ as in Eq.~\eqref{eq:covar}. Assume $\mathcal{L}$ and $\ell_i(\vec{\theta})$ are Lipschitz continuous with at most linear growth, and possess sufficiently smooth derivatives in $\mathcal{G}$. Further assume the physics-guided multiplier $\rho(I)$ is bounded and Lipschitz continuous in $I$. Then:

\noindent\textbf{(i) Vanilla SGD:}~\citep{li2017stochasticmodifiedequationsadaptive} The SDE $d\vec{\theta}_t = -\nabla \mathcal{L}(\vec{\theta}_t) \, dt + \matrix{\Sigma}^{1/2}(\vec{\theta}_t) \, d\vec{w}_t$ is an order-1 weak approximation of SGD, where $t = k \cdot \eta$ is iteration time.

\noindent\textbf{(ii) PGPO:} The modified SDE $d\vec{\theta}_t = -\rho(I_t) \cdot \nabla \mathcal{L}(\vec{\theta}_t) \, dt + \matrix{\Sigma}^{1/2}(\vec{\theta}_t) \, d\vec{w}_t$ is an order-1 weak approximation of PGPO, where $I_t$ is the information measure at time $t$ and the diffusion term remains unchanged.

\noindent \textbf{Proof sketch for (ii).} Since $\rho$ is bounded and Lipschitz in $I$, the modified drift $-\rho(I_t)\nabla\mathcal{L}(\theta_t)$ inherits the Lipschitz and linear-growth conditions of $-\nabla\mathcal{L}$. The result follows by applying~\citep{li2017stochasticmodifiedequationsadaptive} to the modified drift, with the diffusion coefficient unchanged.

\end{theorem}

\noindent Theorem~\ref{thm:weak_convergence} states that the discrete PGPO updates are tracked, in the weak sense, by the modulated-drift SDE as $\eta \to 0$; it does not by itself imply that PGPO converges faster or to a better stationary point than SDPO. We use it solely to confirm that information-based step-size modulation does not break the SDE picture motivating the design. We also note that the form $\rho(I_k) = \min(\exp(\alpha I_k), \rho_{\max})$ is a design choice: any bounded, Lipschitz function of $I_k$ would yield the same conclusion, and we treat $\alpha$ and $\rho_{\max}$ as ordinary hyperparameters.

\newpage
\section{Experiment Details}
\label{sec:experiment_details}

\begin{table}[h]
\centering
\begin{tabular}{lcc}
\toprule
\textbf{Method} & \textbf{SDPO} & \textbf{PGPO} \\
\midrule
\multicolumn{3}{l}{\textbf{General}} \\
Model & Qwen/Qwen3-8B & Qwen/Qwen3-8B \\
Thinking & False & False \\
\midrule
\multicolumn{3}{l}{\textbf{Data}} \\
Max. prompt length & 2048 & 2048 \\
Max. response length & 8192 & 8192 \\
\midrule
\multicolumn{3}{l}{\textbf{Batching}} \\
Question batch size & 32 & 32 \\
Mini batch size & 32 & 32 \\
Number of rollouts & 8 & 8 \\
\midrule
\multicolumn{3}{l}{\textbf{Rollout}} \\
Inference engine & vllm & vllm \\
Temperature & 1.0 & 1.0 \\
\midrule
\multicolumn{3}{l}{\textbf{Validation}} \\
Number of rollouts & 16 & 16 \\
Temperature & 0.6 & 0.6 \\
Top-p & 0.95 & 0.95 \\
\midrule
\multicolumn{3}{l}{\textbf{Training}} \\
Optimizer & AdamW & AdamW \\
Learning rate & $1 \times 10^{-5}$ (constant) & $1 \times 10^{-5}$ (constant) \\
Warmup steps & 10 & 10 \\
Weight decay & 0.01 & 0.01 \\
Gradient Clip Norm & 1.0 & 1.0 \\
Sensitivity Parameter & / & $\{0.5, 1.0, 1.5\}$ \\
\bottomrule
\end{tabular}
\caption{Configuration Comparison}
\end{table}

\begin{table}[h!]
\centering
\small
\label{tab:prompts}
\begin{tabular}{p{3cm} p{10cm}}
\toprule
\textbf{Setting} & \textbf{Prompt Template} \\
\midrule
Unguided & \texttt{\{question\}} \newline Please reason step by step, and put your final answer within \textbackslash boxed\{\}. \\
\midrule
Solution-guided & \texttt{\{question\}} \newline Please reason step by step, and put your final answer within \textbackslash boxed\{\}. \newline Correct solution: \texttt{\{solution\}} \newline Correctly solve the original question. \\
\bottomrule
\end{tabular}
\caption{Prompts for unguided and solution-guided generation following~\citet{hübotter2026reinforcementlearningselfdistillation}.}
\end{table}

\section{Ablation Studies}
\label{sec:ablations}

\begin{table}[ht]
\centering
\label{tab:ablation_alpha}
\begin{tabular}{lcccc}
\toprule
Method & Chemistry & Physics & Biology & Materials \\
\midrule
SDPO & 72.57 & 77.05 & 61.62 & 74.14 \\
\midrule
PGPO with $\alpha=0.5$ & 71.88 & 75.51 & 52.00 & 75.27 \\
PGPO with $\alpha=1.0$ & 76.02 & 76.93 & 56.50 & 77.73 \\
PGPO with $\alpha=1.5$ & 70.94 & 77.56 & 61.25 & 78.65 \\
\bottomrule
\end{tabular}
\caption{Performance comparison between SDPO and PGPO across scientific domains with different sensitivity parameter $\alpha$ settings.}
\end{table}

\begin{figure}[h]
    \centering
    \includegraphics[width=0.9\linewidth]{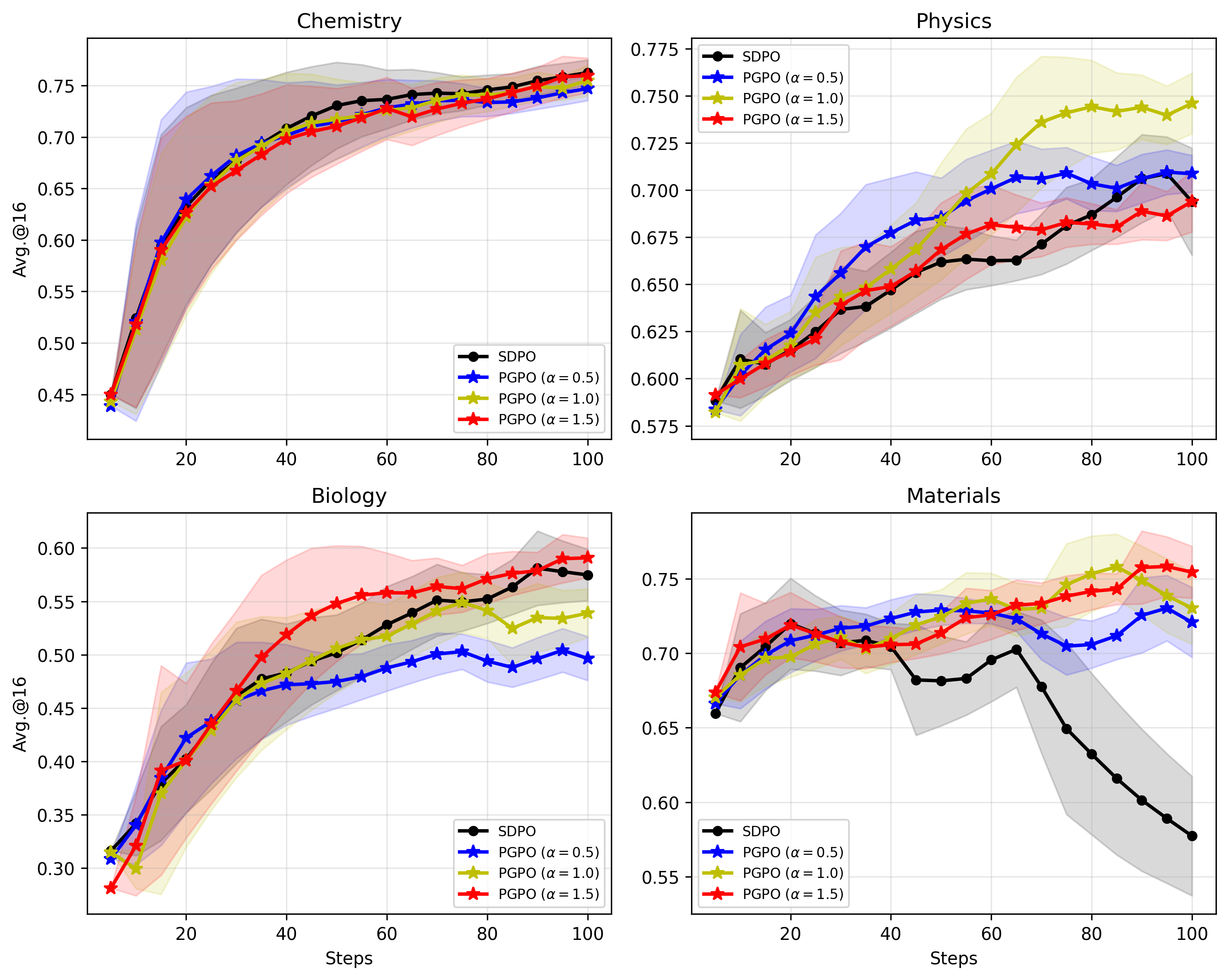}
    \caption{Full ablation study of Physics-Guided Policy Optimization (PGPO) vs. Self-Distillation Policy Optimization (SDPO). Avg.@16 comparison across Science Q\&A tasks: Chemistry, Physics, Biology, and Materials.}

    \label{fig:PGPO-appendix}
\end{figure}

\end{document}